\RequirePackage{fix-cm}
\documentclass{style/hku_mmlab}
\usepackage{amsmath,amssymb}
\usepackage{tabularx}
\usepackage{afterpage}
\usepackage{placeins}
\usepackage{pgfplots}
\pgfplotsset{compat=1.18}
\definecolor{realonewm}{HTML}{426F9D}
\definecolor{realbaseline}{HTML}{B7824E}

\newcommand{\model}{OneWM}
\newcommand{\wammask}{\mbox{WAM\textsubscript{mask}}}
\newcommand{\wamnoisy}{\mbox{WAM\textsubscript{noisy}}}
\newcommand{\wmbaseline}{\mbox{WM}}
\newcommand{\wmadaln}{\mbox{WM\textsubscript{AdaLN}}}

\newcommand{\researchquestion}[1]{\textcolor{sdblue!70!black}{\textit{#1}}}

\title{Can World and Action Modeling Benefit Each Other under Unified Causal Training?}
\renewcommand{\authorfont}{\fontsize{10.5}{13}\selectfont}
\renewcommand{\authorformat}[2][]{{\authorfont\fontfamily{bytesansmedium}\fontseries{m}\selectfont\mbox{#2$^{#1}$}}}
\author[1,2,6,*]{Yue Su}
\author[2,4,*]{Bokai Lin}
\author[2,4,*]{Xinkai Wang}
\author[3]{Chubin Zhang}
\author[5]{Xiangnan Wu}
\author[3]{Kuan Zhang}
\let\onewmAuthorRowOne\authorlist
\renewcommand{\authorlist}{}
\author[1]{Weiyang Jin}
\author[1]{Jiwen Yu}
\author[3]{Yiming Li}
\author[2,4]{Cewu Lu}
\author[2,4,\dagger]{Lixin Yang}
\author[1,\dagger]{Xihui Liu}
\let\onewmAuthorRowTwo\authorlist
\renewcommand{\authorlist}{%
  \onewmAuthorRowOne\\[1.5pt]
  \onewmAuthorRowTwo}
\renewcommand{\affiliationfont}{\fontsize{8.75}{10.5}\selectfont}
\renewcommand{\affiliationformat}[2][]{{\affiliationfont\fontfamily{phv}\fontseries{m}\selectfont$^{#1}$#2}}
\affiliation[1]{MMLab@HKU}
\affiliation[2]{MVIG@SJTU}
\affiliation[3]{THU}
\affiliation[4]{SII}
\affiliation[5]{UCAS}
\affiliation[6]{\IAAITeam}
\date{\today}
\checkdata[Project Page]{\url{https://selen-suyue.github.io/OneWMNet}}

\abstract{
World and action modeling are closely related, yet their reciprocal
benefits under unified causal training
remain unclear. We introduce \textbf{\model{}}, a simple framework for
unified causal training that treats actions as both conditions for world
prediction and targets for policy learning within a shared video--action
architecture. Using this framework, we investigate whether learning
forward dynamics conditioned on specified actions improves action
learning, and whether learning to generate actions in turn improves
action-conditioned world modeling.
We use LIBERO, LIBERO-Plus, and RoboCasa365 as embodied testbeds to
evaluate both policy learning and world modeling, complemented by
real-world policy experiments. Learning the visual consequences of
specified actions improves policy performance and generalization across
changing scenes. In turn, action learning improves multi-view world-prediction fidelity
under the same architecture. Together, these results answer both questions
positively, supporting reciprocal benefits between world modeling
and action generation under unified causal training.
}

\begin{document}
\maketitle

% A single compact first-page footnote explains author symbols.
\begingroup
  \renewcommand{\thefootnote}{}
  \begin{NoHyper}
    \footnotetext{\fontsize{8}{10}\selectfont
      $^{*}$Equal contribution.\quad $^{\dagger}$Corresponding authors.}
  \end{NoHyper}
\endgroup

% Schedule Figure 1 after the first page ships out, at the top of page 2.
% Keep this registration here rather than moving it beside its first mention.
\afterpage{%
  \begin{figure}[!t]
    \centering
    \includegraphics[width=\textwidth]{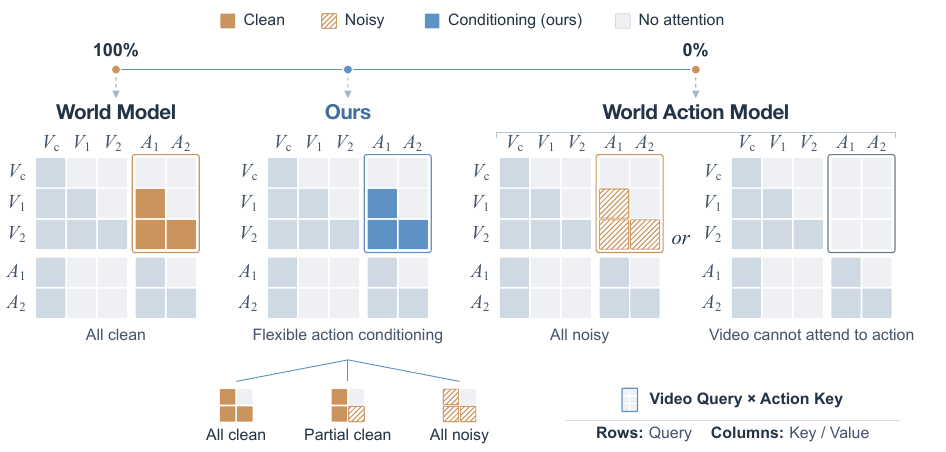}
    \caption{\CaptionLead{Action conditioning for world and action modeling.}
    Rows denote queries and columns denote keys/values.
    $V_c$, $V_{1,2}$, and $A_{1,2}$ represent visual context,
    future video, and aligned actions, respectively.
    Outlined blocks compare world prediction conditioned on clean ground-truth actions
    (solid orange) with WAM prediction using noisy actions (hatching)
    or masked action keys (light gray).
    Percentages indicate the fraction of clean ground-truth actions.
    \model{} (blue) mixes all-clean, partial-clean, and all-noisy examples
    to examine whether world and action modeling benefit each other
    under unified causal training.}
    \label{fig:action-conditioning}
  \end{figure}%
}

\section{Introduction}
Predicting how actions affect the physical world and learning to generate
actions are closely related problems~\citep{nvidia2025cosmos}.
World models predict future observations from specified
actions~\citep{zhu2025uwm,quevedo2025worldgym}, while world-action models
(WAMs) use video generation to guide action
generation~\citep{liang2025videopolicy,ye2026dreamzero}.
These complementary objectives motivate studying world and action
modeling within a unified framework.

% Video foundation models support policy learning by predicting future
% observations~\citep{liang2025videopolicy,shen2025videovla,ye2026dreamzero}.
% One line of work~\citep{yuan2026fastwam,bi2025motus,fan2026aim} adopts
% mixture-of-transformers (MoT) architectures~\citep{liang2025mixture} that
% allow action policy to attend to bidirectional video
% representations~\citep{yang2024cogvideox,kong2024hunyuanvideo,nvidia2025cosmos,wan2025wan}.
% These representations provide effective policy priors, leading to strong
% manipulation performance.
% Other work~\citep{li2025uva,zhu2025uwm,bi2025motus,nvidia2026cosmos3}
% unifies action and video generation within shared architectures,
% demonstrating the benefits of video generation for action learning.
% In parallel, causal WAMs~\citep{li2026lingbotva,ye2026gigaworldpolicy,ye2026dreamzero}
% couple action and video generation through directional attention patterns,
% aiming to learn more physically consistent representations of causal dynamics.

Video foundation models support policy learning by predicting future
observations~\citep{liang2025videopolicy,shen2025videovla,ye2026dreamzero}.
One line of work~\citep{yuan2026fastwam,bi2025motus,fan2026aim} uses
mixture-of-transformers (MoT) architectures~\citep{liang2025mixture},
in which action tokens attend to bidirectional video
representations~\citep{yang2024cogvideox,kong2024hunyuanvideo,nvidia2025cosmos,wan2025wan}.
These visual representations serve as effective policy priors and
support strong manipulation performance.
Related studies~\citep{li2025uva,zhu2025uwm,bi2025motus,nvidia2026cosmos3}
jointly model video and action generation within shared architectures,
further demonstrating how video generation benefits action learning.
In parallel, causal WAMs~\citep{li2026lingbotva,ye2026gigaworldpolicy,ye2026dreamzero}
use directional attention to couple video and action generation,
aiming to capture causal dynamics in a more physically consistent way.

{\emergencystretch=1em
These results demonstrate the value of video prediction for action learning,
yet the interaction between action-conditioned world modeling and action
generation under temporal causality warrants further study.
With bidirectional backbones augmented with an action-inference branch,
the video branch often retains its pretrained
attention~\citep{yuan2026fastwam}, reflecting the premise that existing
visual representations suffice as policy priors.
Causal WAMs allow video tokens to attend to actions, but these actions are
noisy variables under joint denoising~\citep{ye2026gigaworldpolicy,ye2026dreamzero}.
Joint generation therefore differs from forward dynamics prediction
conditioned on specified clean actions.
Conditioning video generation on ground-truth actions makes forward dynamics
an explicit learning objective, raising our first question:
\researchquestion{Can learning such action-conditioned forward dynamics
improve action learning within the same causal model?}
Within the same framework, action generation may in turn provide a learning
signal for visual prediction, motivating the converse question:
\researchquestion{Can learning to generate actions, in turn, improve
action-conditioned world modeling within the same causal model?}
\par}

% Existing architectures provide a straightforward basis for unified causal
% training, allowing us to investigate these questions within a shared model.
% In a causal MoT framework, Fig.~\ref{fig:action-conditioning} illustrates
% two endpoints: an \emph{all-clean} setting, where a world model predicts
% video conditioned on clean ground-truth actions, and an \emph{all-noisy}
% setting, where all action tokens are noisy and serve as prediction targets
% for a WAM.
% To study these two settings jointly, we consider a unified causal model,
% termed \textbf{\model{}}, that accommodates both all-clean and all-noisy
% action conditioning.
% Beyond these endpoints, \model{} also supports a \emph{partial-clean}
% setting, where an action sequence consists of a clean prefix and a noisy
% suffix.
% Here, the action branch denoises the suffix conditioned on the clean prefix,
% while the video branch predicts future observations with only partial
% clean-action conditioning.
% Each training batch mixes samples from these three settings, enabling joint
% world and action modeling within the same causal model.

Existing architectures provide a straightforward basis for unified causal
training, allowing us to investigate these questions within a shared model.
In a causal MoT framework, Fig.~\ref{fig:action-conditioning} illustrates
two endpoints: an \emph{all-clean} setting, where a world model predicts
video conditioned on clean ground-truth actions, and an \emph{all-noisy}
setting, where noisy action tokens serve as generation targets while
conditioning video prediction.
The figure also shows an representative WAM formulation that prevents
video tokens from attending to action tokens.
Both WAM formulations are organized around action generation through
denoising, limiting action-to-video information flow: the video branch
receives only noisy actions or no action information at all.
To connect explicit forward-dynamics learning with action generation,
we consider \textbf{\model{}}, a unified causal model that supports
both all-clean and all-noisy conditioning, together with a
\emph{partial-clean} setting, which uses a clean action prefix to condition both
future-video prediction and denoising of the remaining noisy suffix.
Each training batch mixes the three settings, enabling us to examine
reciprocal benefits between forward-dynamics learning and action
generation within the same causal model.

We conduct extensive experiments across multiple simulated and real-world
environments to examine whether world and action modeling benefit each
other under unified causal training. Our results provide positive answers to both questions.
First, under \model{}'s unified causal training, incorporating
clean-action conditioning into world model learning substantially improves
the policy's action-prediction performance, outperforming the evaluated
WAMs with action-masked or noisy-action-conditioned video branches.
Second, learning to generate actions also improves action-conditioned
world modeling, showing that the benefits can be reciprocal in this
setting. We further analyze the interaction between the two learning tasks.

\section{Related Work}

\subsection{World Action Models}
World-action models bring predictive world representations into policy
learning. One line of work uses generated videos as visual plans or
synthetic demonstrations, recovering executable actions through inverse
dynamics or goal-conditioned
controllers~\citep{du2023unipi,du2023vlp,jang2025dreamgen}.
Other approaches integrate visual forecasting into the policy itself,
using predictive video
features~\citep{hu2024vpp,pai2025mimicvideo,liao2025genie} or jointly
learning future observations and
actions~\citep{wu2023gr1,liang2025videopolicy,shen2025videovla,kim2026cosmospolicy,ye2026dreamzero,fan2026aim,wang2026openwam}.
These formulations use temporal supervision and video pretraining to
improve policy generalization and data efficiency.

Recent work also separates predictive learning from the cost of future
synthesis. Efficient WAMs retain video co-training while reducing or
bypassing future-video generation at
deployment~\citep{yuan2026fastwam,li2026lightwam}.
Latent and condition-space approaches couple future-representation
prediction with action
generation~\citep{su2026worldguidance,lin2026jepawam,du2026actualizer}, while representation
distillation transfers world-model features to policies without adding a
generative process at inference~\citep{dao2026thaw}.
We build on these predictive approaches by making the visual consequences
of specified actions an explicit learning target during policy training.

\subsection{World Models}
World models learn transition dynamics for planning and policy
optimization, either in compact latent
spaces~\citep{ha2018worldmodels,hafner2023dreamer,assran2025vjepa2} or
through visual forecasting~\citep{wu2024ivideogpt,huang2025vid2world}.
Large-scale video pretraining provides reusable visual dynamics
priors~\citep{nvidia2025cosmos,wan2025wan}, while robotic world models
develop control interfaces that relate these priors to concrete robot
commands. Frame-level action injection and spatially grounded kinematic
conditioning support fine-grained, controllable
rollouts~\citep{zhu2024irasim,guo2025ctrlworld,wu2026oscar}.
Latent-action pretraining offers a complementary route to learning
interaction dynamics from unlabeled videos before adapting to robot
controls~\citep{gao2025adaworld,gao2026dreamdojo}.

These predictors increasingly serve as learned simulators for policy
evaluation and improvement, supplying imagined rollouts and synthetic
experience~\citep{quevedo2025worldgym,guo2025ctrlworld,qiu2026gesim,jiang2025world4rl}.
Their dynamics priors can also be transferred directly to policy
learning~\citep{huang2026a2world}.
These applications require predictions that follow the supplied actions,
a property distinguished from visual plausibility in recent
evaluations~\citep{chen2026worldecho}.
Our work also treats actions as learning targets, using action-generation
supervision to strengthen the representations underlying these predictions.

\subsection{Unified World and Action Modeling}
Unified models support world prediction and action generation through
shared representations and flexible
conditioning~\citep{li2025uva,zhu2025uwm,bi2025motus,nvidia2026cosmos3}.
This line spans bidirectional diffusion
frameworks~\citep{shen2025videovla,nvidia2026cosmos3} and causal designs
with autoregressive or directional information
flow~\citep{li2026lingbotva,ye2026dreamzero,ye2026gigaworldpolicy}.

Existing studies report both reciprocal benefits and task trade-offs.
For example, WorldVLA~\citep{cen2025worldvla} interleaves discrete image and action
tokens in a shared autoregressive model, coupling action generation with
action-conditioned next-image prediction.
Joint training improves both action generation and long-horizon visual
prediction.
By contrast, Cosmos~3~\citep{nvidia2026cosmos3} reports that joint
learning with video improves action prediction under bidirectional
video--action diffusion, but co-training with action prediction leads to
a small decline in video-generation fidelity.

These findings arise from different modeling settings: WorldVLA
demonstrates mutual benefits through discrete autoregressive image--action
modeling, whereas Cosmos~3 studies joint video--action learning with
bidirectional attention.
Together, they leave open how temporal causality shapes the learning
of physical dynamics and the resulting benefits for video and action
prediction.
We investigate this interaction within a shared causal video--action
MoT framework, explicitly coupling action-conditioned forward-dynamics
learning with action generation to examine their reciprocal benefits.

\section{Method}

\begin{figure}[!htbp]
  \centering
  \includegraphics[width=0.95\textwidth]{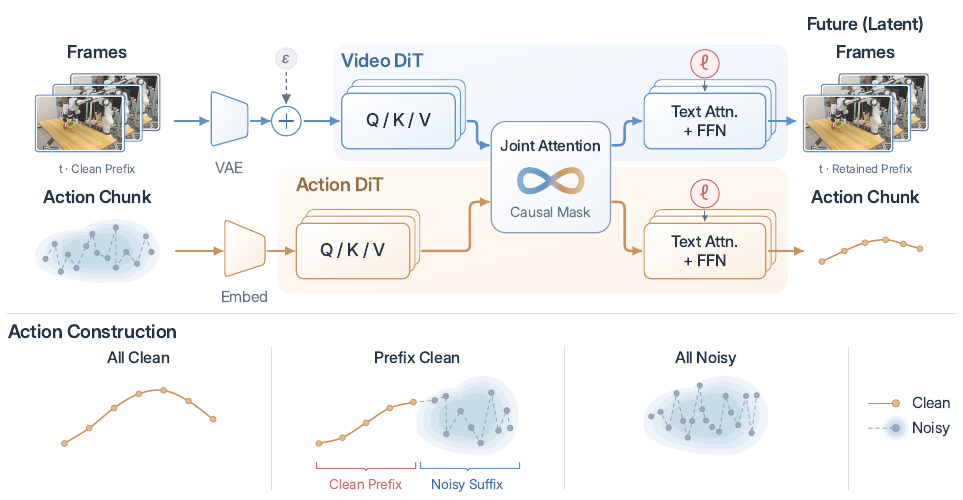}
  \caption{\CaptionLead{Overview of \model{}.}
  A VAE encodes frames into latents; the observed prefix remains clean,
  and the video expert denoises future latents. Action chunks are embedded for the action
  expert. Causal joint attention couples the two experts, each receiving
  instruction $\ell$ through its own text cross-attention
  (Sec.~\ref{sec:causal-modeling}). The lower panels illustrate all-clean,
  clean-prefix/noisy-suffix, and all-noisy action construction, unifying
  world prediction and action generation (Sec.~\ref{sec:action-conditioning}).}
  \label{fig:method-pipeline}
\end{figure}

\subsection{Problem Formulation}
We consider future-video prediction and action generation from an observed
visual context and a language instruction. Let $\mathbf{o}^{\mathrm{ctx}}$
denote the observed frame prefix, $\mathbf{o}^{\mathrm{fut}}$ the future
frames, and $\mathbf{a}=(a_1,\ldots,a_K)$ an action chunk over the prediction
horizon. We denote the language instruction by $\ell$.
As illustrated in Fig.~\ref{fig:method-pipeline}, a video VAE encodes frames
into latents $\mathbf{z}^{\mathrm{ctx}}$ and $\mathbf{z}^{\mathrm{fut}}$,
while the action chunk is represented by action embeddings. The observed
video prefix remains clean, and future video latents are denoising targets.

World modeling predicts future video conditioned on specified clean
actions, while action modeling generates an action chunk conditioned on
the visual context and instruction. With $\mathcal{C}=(\mathbf{z}^{\mathrm{ctx}},\ell)$
and model parameters $\theta$, these tasks correspond to
\begin{equation}
  p_{\theta}\!\left(\mathbf{z}^{\mathrm{fut}}\mid\mathcal{C},\mathbf{a}\right)
  \quad\text{and}\quad
  p_{\theta}\!\left(\mathbf{a}\mid\mathcal{C}\right).
  \label{eq:prediction-tasks}
\end{equation}

\model{} implements both tasks with separate video and action experts
that exchange information through temporally causal joint attention.
Each expert receives language conditioning through its own text
cross-attention. Across training samples, actions serve as clean
conditions, noisy prediction targets, or a clean prefix followed by a
noisy suffix. This formulation allows us to study the interaction between
clean-action-conditioned world prediction and action learning within the
same causal model.

\subsection{Temporally Causal Video--Action Modeling}
\label{sec:causal-modeling}
To examine the interaction between forward-dynamics learning and action
learning, \model{} couples the two tasks through a common causal MoT
backbone~\citep{liang2025mixture} (Fig.~\ref{fig:method-pipeline}).
We initialize the video expert from pretrained
Wan2.2-TI2V-5B~\citep{wan2025wan} and reuse its video VAE and
text encoder.
The video expert models visual dynamics in latent space, and the action
expert models action chunks. Modality-specific attention projections and
feed-forward layers accommodate their different representations. Both
experts receive the instruction $\ell$ through separate text
cross-attention modules, allowing the same task semantics to condition
visual prediction and action generation.

The causal structure is specified by the temporal alignment of video and
action blocks. Let $\mathcal{V}_0$ denote the observed visual context,
$\mathcal{V}_i$ the tokens of future latent frame $i$, and
$\mathcal{A}_i$ the action segment aligned with that frame. For queries in
block $i\geq 1$, the sets of visible key tokens are
\begin{equation}
  \begin{aligned}
    \mathcal{S}_v(i) &= \mathcal{V}_{\leq i}\cup\mathcal{A}_{\leq i},\\
    \mathcal{S}_a(i) &= \mathcal{V}_0\cup\mathcal{A}_{\leq i},
  \end{aligned}
  \label{eq:causal-visibility}
\end{equation}
where $\mathcal{V}_{\leq i}=\bigcup_{j=0}^{i}\mathcal{V}_j$ and
$\mathcal{A}_{\leq i}=\bigcup_{j=1}^{i}\mathcal{A}_j$. Thus, future video
queries access the video and action prefixes up to their aligned block,
while action queries use the observed visual context and the
corresponding action prefix. Context queries remain confined to the
observation stream. Tokens within a future frame or an action segment
can interact freely, and dependencies between blocks follow temporal
order.

Following MoT, the experts use their own query, key, and value projections
while performing joint attention over the permitted tokens. Writing
$Q_m$, $K_m$, and $V_m$ for expert $m\in\{v,a\}$, the attention
aggregation is
\begin{equation}
  U_m = \operatorname{softmax}_{\mathcal{S}_m}\!\left(
    \frac{Q_m K^{\top}}{\sqrt{d}}
  \right)V,
  \qquad
  K=\begin{bmatrix}K_v\\K_a\end{bmatrix},\quad
  V=\begin{bmatrix}V_v\\V_a\end{bmatrix},
  \label{eq:causal-joint-attention}
\end{equation}
where $d$ is the attention-head dimension and
$\operatorname{softmax}_{\mathcal{S}_m}$ normalizes each query row over
its visible keys only. The aggregated features $U_m$ are then processed
by the corresponding expert's output projection and remaining layers.

This asymmetric coupling connects the two learning tasks under temporal
causality. Attention to action features grounds forward-dynamics
prediction in control signals, while attention to the observed video
features supports action generation. Both objectives train the
interconnected experts, providing routes for learning signals from one
task to shape representations used by the other. Across the training
regimes described next, we retain these visibility relations and vary
which actions serve as clean conditions or prediction targets. This
allows us to study the interaction between the two learning objectives
within the same causal architecture.

\subsection{Unified Action Conditioning}
\label{sec:action-conditioning}
Within the causal backbone above, we vary the role of actions between
specified conditions for world prediction and targets for action
generation. As illustrated by the Action Construction panels in
Fig.~\ref{fig:method-pipeline}, these roles are unified by the length
$r\in\{0,\ldots,K\}$ of a clean action prefix. Writing
$\widetilde{a}_k(\tau)$ for action $a_k$ corrupted at noise level $\tau$,
the action input is
\begin{equation}
  a_k^{(r,\tau)} =
  \begin{cases}
    a_k, & k\leq r,\\
    \widetilde{a}_k(\tau), & k>r.
  \end{cases}
  \label{eq:action-conditioning}
\end{equation}
The action expert receives a noise-level embedding for each token:
zero for clean-prefix tokens ($k\leq r$) and the same sampled level
$\tau$ for all noisy-suffix tokens ($k>r$). These embeddings distinguish
specified conditions from denoising targets within the action sequence.
All regimes retain the visibility relations in
Eq.~\ref{eq:causal-visibility}. The observed video prefix remains clean,
and future video latents are denoised throughout.

\paragraph{All-clean ($r=K$).}
All actions are clean conditions for future video prediction. This regime
learns action-conditioned forward dynamics, with no action denoising loss.

\paragraph{Partial-clean ($0<r<K$).}
A clean prefix conditions denoising of the remaining action suffix,
corresponding to the Prefix Clean panel in
Fig.~\ref{fig:method-pipeline}. Within this regime, we choose between two sampling strategies
according to a fixed mixing ratio. The first uniformly selects
a boundary between the video-aligned action blocks $\mathcal{A}_i$
and perturbs it by up to two action steps to obtain $r$, clipped
to $\{1,\ldots,K-1\}$. The second samples $r$ uniformly from this range. This favors temporally aligned conditions
while retaining varied prefix lengths. Only the noisy suffix contributes
to the action loss, while video prediction uses both clean and noisy
action features.

\paragraph{All-noisy ($r=0$).}
All actions are denoising targets, predicted from the observed visual
context and instruction. Future video prediction is jointly trained with
noisy action features as conditions.

We sample the regime and clean-prefix length per example, mixing these
forms of conditioning within each batch. The same experts thus learn
both specified-action dynamics and action generation, allowing us to
assess their reciprocal benefits under a fixed causal architecture.

\subsection{Joint Training Objective}
\label{sec:joint-objective}
Building on the conditioning regimes in
Sec.~\ref{sec:action-conditioning}, we jointly train the video and action
experts with flow matching~\citep{lipman2022flow}. For a clean target
$\mathbf{x}$, the noisy input is
$\mathbf{x}(\tau)=(1-\tau)\mathbf{x}+\tau\boldsymbol{\epsilon}$,
where $\boldsymbol{\epsilon}\sim\mathcal{N}(0,I)$, and the target velocity
is $\boldsymbol{\epsilon}-\mathbf{x}$. Video and action noise levels
$\tau_v$ and $\tau_a$ are sampled independently. The clean visual context
and action prefix remain conditioning inputs.

Let $\widehat{\mathbf{u}}_v$ and $\widehat{\mathbf{u}}_a$ denote velocity
predictions for future video latents and the action chunk, respectively.
With the clean-prefix length $r$ defined in
Eq.~\ref{eq:action-conditioning}, the joint objective is
\begin{equation}
  \begin{aligned}
    \mathcal{L}_v &= \mathbb{E}\!\left[
      w_v(\tau_v)\left\|
      \widehat{\mathbf{u}}_v-
      (\boldsymbol{\epsilon}_v-\mathbf{z}^{\mathrm{fut}})
      \right\|_{\mathrm{mean}}^2\right],\\
    \mathcal{L}_a &= \mathbb{E}_{r<K}\!\left[
      w_a(\tau_a)\left\|
      \widehat{\mathbf{u}}_{a,>r}-
      (\boldsymbol{\epsilon}_{a,>r}-\mathbf{a}_{>r})
      \right\|_{\mathrm{mean}}^2\right],\\
    \mathcal{L} &= \lambda_v\mathcal{L}_v+
      \lambda_a\mathcal{L}_a.
  \end{aligned}
  \label{eq:joint-objective}
\end{equation}
Here $\|\cdot\|_{\mathrm{mean}}^2$ averages squared error over valid target
elements, $w_v$ and $w_a$ are noise-level weights, and $\lambda_v$ and
$\lambda_a$ balance the two objectives. The expectations cover training
examples, conditioning regimes, and noise. The action loss averages only
examples with a noisy suffix ($r<K$), and is zero when no such targets
are present. Thus, all-clean examples provide forward-dynamics
supervision, while partial-clean and all-noisy examples additionally
supervise action generation. Both losses optimize the coupled experts
through causal joint attention.

\subsection{Inference}
\label{sec:inference}
We use two inference settings corresponding to the all-clean and
all-noisy endpoints in Sec.~\ref{sec:action-conditioning}. Sampling
integrates the velocity fields learned in Sec.~\ref{sec:joint-objective}
from noise ($\tau=1$) to clean outputs ($\tau=0$), with the observed visual
context and language instruction held fixed.

\paragraph{World prediction.}
The action chunk is supplied as clean conditioning ($r=K$) and
remains fixed throughout sampling. The video expert denoises future
video latents initialized from Gaussian noise, conditioned on
$\mathcal{C}=(\mathbf{z}^{\mathrm{ctx}},\ell)$---the observed visual context
and language instruction---together with the specified actions. The video VAE decodes the resulting
latents into predicted frames.

\paragraph{Action generation.}
The action chunk is initialized from Gaussian noise ($r=0$)
and denoised by the action expert, conditioned on the same observed
visual context and language instruction.
Observed video representations are reused across denoising steps.
For closed-loop control, a new action chunk is generated from the
latest observations at each replanning step.

\section{Experiments}
\begin{figure*}[!ht]
  \centering
  \includegraphics[width=\textwidth]{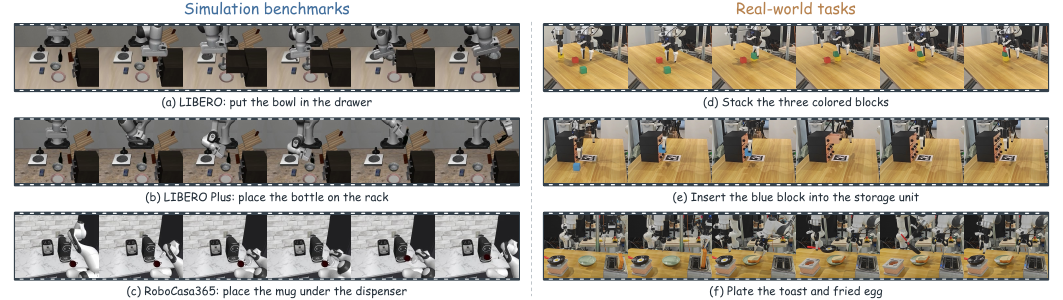}
  \caption{\CaptionLead{Evaluation tasks and environments.} We evaluate \model{}
  on LIBERO, LIBERO-Plus, and RoboCasa365 in simulation, and on three real-world
  tasks: block stacking, block insertion, and breakfast arrangement.}
  \label{fig:experiment-episodes}
\end{figure*}

We evaluate how world and action modeling affect each other under unified
causal training.
Figure~\ref{fig:experiment-episodes} summarizes the simulation benchmarks
and real-world tasks used in our study.

\subsection{Experimental Setup}
\label{sec:experimental-setup}
\paragraph{Benchmarks and evaluation.}
We evaluate on three simulation benchmarks: LIBERO~\citep{liu2023libero},
LIBERO-Plus~\citep{fei2025liberoplus} for generalization under diverse
perturbations, and RoboCasa365~\citep{nasiriany2026robocasa365} for mobile
manipulation in complex environments. Real-world experiments cover block
stacking, block insertion, and breakfast arrangement
(Fig.~\ref{fig:experiment-episodes}). We report task success rates for
policy evaluation and PSNR, SSIM, LPIPS, FID, and FVD for world prediction.
Detailed training and evaluation protocols are provided in
Appendix~\ref{app:evaluation-protocol}.

\paragraph{Comparisons.}
To answer our first question---whether learning action-conditioned
forward dynamics improves action learning---we compare \model{} with
the two WAM variants illustrated in Fig.~\ref{fig:action-conditioning}.
The \underline{action-masked WAM} (\wammask{}) prevents future-video
tokens from attending to action tokens, blocking action-to-video
information flow.
The \underline{noisy-action-conditioned WAM} (\wamnoisy{}) allows
such attention but conditions video prediction on noisy actions
throughout training.
Both jointly learn video and action prediction; \model{} additionally
trains with clean and partially clean actions.
To answer our second question---whether learning to generate actions
improves action-conditioned world modeling---we compare \model{}
with the \underline{world-model baseline} (\wmbaseline{}).
Corresponding to the all-clean setting in
Fig.~\ref{fig:action-conditioning}, \wmbaseline{} predicts future video
conditioned on clean actions without an action-prediction objective.

\subsection{World Modeling for Action Learning}
\label{sec:world-for-action}
To address our first question, we compare the policies learned by
\model{} and the two WAM variants under the same causal architecture.
All three jointly learn video and action prediction, while \model{}
also learns forward dynamics conditioned on clean actions. This comparison
tests whether explicitly learning the visual consequences of specified
actions benefits action learning.

\begin{table*}[!t]
  \centering
  \captionsetup{skip=5pt}
    \caption{\CaptionLead{Policy performance in simulation.}
    Success rates (\%) on LIBERO, LIBERO-Plus, and RoboCasa365.
    Bold highlights the best result among internal variants.
    Bkgd., Lang., and Init.\ denote background, language, and robot
    initialization, respectively; S/U and Comp.\ denote seen/unseen
    and composite tasks in RoboCasa365.}
  \label{tab:simulation-policy}
  \small
  \renewcommand{\arraystretch}{1.06}
  \setlength{\aboverulesep}{1.8pt}
  \setlength{\belowrulesep}{1.8pt}
  \begin{minipage}[t]{0.485\textwidth}
  \centering
  \setlength{\tabcolsep}{1.5pt}
  \begin{tabularx}{\linewidth}{@{}p{0.27\linewidth}*{5}{>{\centering\arraybackslash}X}@{}}
    \toprule
    \multicolumn{6}{c}{\sdtablehead{LIBERO}} \\
    \midrule
    Method & Spatial & Object & Goal & Long & Avg. \\
    \midrule
    $\pi_0$~\citep{black2024pi0} & 98.00 & 96.80 & 94.40 & 88.40 & 94.40 \\
    WorldVLA~\citep{cen2025worldvla} & 87.60 & 96.20 & 83.40 & 60.00 & 81.80 \\
    Fast-WAM~\citep{yuan2026fastwam} & 98.20 & 100.00 & 97.00 & 95.20 & 97.60 \\
    \midrule
    \wammask{} & 93.80 & 96.20 & \textbf{96.80} & 83.00 & 92.45 \\
    \wamnoisy{} & 94.00 & 95.80 & 95.60 & 85.00 & 92.60 \\
    \textbf{\model{}} & \textbf{96.60} & \textbf{98.60} & \textbf{96.80} & \textbf{87.80} & \textbf{94.95} \\
    \bottomrule
  \end{tabularx}
  \end{minipage}\hfill
  \begin{minipage}[t]{0.485\textwidth}
  \centering
  \setlength{\tabcolsep}{1.5pt}
  \begin{tabularx}{\linewidth}{@{}p{0.37\linewidth}*{4}{>{\centering\arraybackslash}X}@{}}
    \toprule
    \multicolumn{5}{c}{\sdtablehead{RoboCasa365}} \\
    \midrule
    {\footnotesize Method} & {\footnotesize Atomic-S} & {\footnotesize Comp.-S} & {\footnotesize Comp.-U} & {\footnotesize Overall} \\
    \midrule
    $\pi_0$~\citep{black2024pi0} & 34.60 & 6.10 & 1.10 & 14.80 \\
    $\pi_{0.5}$~\citep{intelligence2025pi05} & 39.60 & 7.10 & 1.20 & 16.90 \\
    GigaWorld-Policy~\citep{ye2026gigaworldpolicy} & 44.40 & 11.80 & 2.90 & 20.70 \\
    GR00T N1.6~\citep{nvidia2025groot16} & 51.10 & 9.40 & 1.70 & 21.90 \\
    \midrule
    \wammask{} & 55.89 & 19.63 & 6.00 & 28.32 \\
    \textbf{\model{}} & \textbf{68.78} & \textbf{30.25} & \textbf{7.00} & \textbf{36.68} \\
    \bottomrule
  \end{tabularx}
  \end{minipage}
  \par\vspace{7pt}
  \begin{minipage}[t]{1.0\textwidth}
  \centering
  \setlength{\tabcolsep}{1.5pt}
  \begin{tabularx}{\linewidth}{@{}p{0.20\linewidth}*{8}{>{\centering\arraybackslash}X}@{}}
    \toprule
    \multicolumn{9}{c}{\sdtablehead{LIBERO-Plus}} \\
    \midrule
    Method & Bkgd. & Camera & Lang. & Light & Layout & Init. & Noise & Overall \\
    \midrule
    $\pi_0$~\citep{black2024pi0} & 81.40 & 13.80 & 58.80 & 85.00 & 68.90 & 6.00 & 79.00 & 53.60 \\
    WorldVLA~\citep{cen2025worldvla} & 17.10 & 0.10 & 41.60 & 43.70 & 38.00 & 27.90 & 10.90 & 25.00 \\
    Fast-WAM~\citep{yuan2026fastwam} & 52.30 & 16.30 & 68.10 & 78.90 & 60.50 & 44.40 & 37.90 & 49.70 \\
    \midrule
    \wammask{} & 54.00 & 11.13 & 73.85 & \textbf{84.24} & 62.36 & 47.68 & 50.41 & 53.37 \\
    \wamnoisy{} & 53.07 & 10.82 & 70.33 & 81.87 & 63.08 & \textbf{49.68} & 48.34 & 52.50 \\
    \textbf{\model{}} & \textbf{57.34} & \textbf{18.39} & \textbf{79.57} & 79.86 & \textbf{71.08} & 47.81 & \textbf{67.08} & \textbf{59.27} \\
    \bottomrule
  \end{tabularx}
  \end{minipage}
\end{table*}

\paragraph{In-distribution performance and generalization.}
\model{} outperforms both WAM variants on in-distribution LIBERO and
out-of-distribution LIBERO-Plus (Table~\ref{tab:simulation-policy}).
On LIBERO, the gains span spatial, object, and long-horizon tasks,
while goal performance matches the stronger variant. On LIBERO-Plus,
the advantages extend to background, camera, language, layout, and
sensor-noise perturbations, with particularly strong gains under sensor
noise. Camera and layout changes require the policy to adapt to different
visual configurations while preserving the intended interaction. The
improvements in these settings suggest that learning action-conditioned
dynamics strengthens representations of task-relevant scene changes.
The benefit remains uneven under lighting changes and robot initialization.

The comparison with Fast-WAM further highlights this generalization
advantage. Although \model{} trails Fast-WAM on LIBERO, it outperforms
it across all LIBERO-Plus perturbation categories and achieves the best
overall result among the listed policies. Fast-WAM~\citep{yuan2026fastwam}
uses bidirectional attention within its video and action experts while
separating future-video prediction from action generation.
In \model{}, clean-action conditioning and causal joint attention
ground video prediction in specified controls and their temporal order.
This supervision encourages the visual expert to capture how actions
change object configurations, shaping the observed representations used
by the action expert. The stronger performance under distribution shift
is consistent with learning transferable interaction dynamics rather
than relying primarily on correlations specific to familiar scenes.

\paragraph{Manipulation in diverse  scenes.}
On RoboCasa365, \model{} surpasses \wammask{} and the listed reference
policies across all task groups. The benchmark spans stationary and mobile
manipulation across diverse scene layouts, object arrangements, and
articulated fixtures~\citep{nasiriany2026robocasa365}. Many tasks involve
base motion, which adds viewpoint changes to the spatial changes caused
by manipulation. Learning action-conditioned forward dynamics encourages
the visual expert to track how controls change these object--scene
relationships. The policy gains are consistent with this supervision
providing richer spatial representations for manipulation in more complex
and changing scenes.

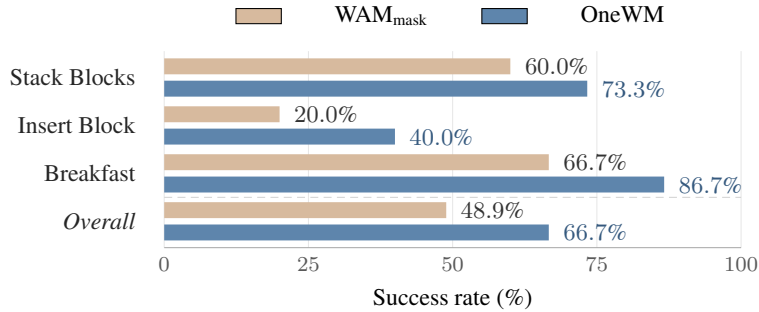
\begin{figure}[!htbp]
  \centering
  \begin{tikzpicture}
    \begin{axis}[
      scale only axis,
      width=0.46\textwidth, height=2.6cm,
      xbar, bar width=6pt,
      xmin=0, xmax=100, ymin=-0.55, ymax=3.55,
      enlarge y limits=false,
      ytick={0,1,2,3},
      yticklabels={\textit{Overall},Breakfast,Insert Block,Stack Blocks},
      xtick={0,25,50,75,100},
      xlabel={Success rate (\%)},
      axis lines*=left,
      axis line style={draw=black!35,line width=0.35pt},
      y axis line style={draw=none},
      tick style={draw=none},
      xmajorgrids=true,
      grid style={draw=black!10,line width=0.3pt},
      xticklabel style={font=\footnotesize,text=black!65},
      yticklabel style={font=\small,text=black!85,xshift=-3pt},
      xlabel style={font=\small},
      legend style={at={(0.5,1.07)},anchor=south,legend columns=2,
        draw=none,fill=none,font=\small,column sep=18pt},
      legend cell align=left,
      area legend,
      nodes near coords={\pgfmathprintnumber[fixed,precision=1,fixed zerofill]{\pgfplotspointmeta}\%},
      nodes near coords align={horizontal},
      every node near coord/.append style={font=\small,text=black!80,xshift=2pt},
      clip=false,
    ]
      \addplot[bar shift=4.2pt,fill=realbaseline!55,draw=none]
        coordinates {(48.88889,0) (66.66667,1) (20,2) (60,3)};
      \addplot[bar shift=-4.2pt,fill=realonewm!85,draw=none,
        every node near coord/.append style={text=realonewm!80!black}]
        coordinates {(66.66667,0) (86.66667,1) (40,2) (73.33333,3)};
      \legend{\wammask{},\model{}}
      \draw[black!18,densely dashed,line width=0.35pt]
        (axis cs:0,0.5) -- (axis cs:100,0.5);
    \end{axis}
  \end{tikzpicture}
  \caption{\CaptionLead{Real-world policy performance.}
  Success rates on three manipulation tasks; Overall aggregates the tasks.}
  \label{fig:real-world-policy}
\end{figure}

\paragraph{Real-world validation.}
Stacking tests relative placement, insertion tests precise alignment, and
breakfast arrangement tests successive manipulations as the scene changes.
\model{} improves over \wammask{} on all three tasks
(Fig.~\ref{fig:real-world-policy}), especially insertion and breakfast
arrangement. Whereas \wammask{} excludes actions from video prediction,
\model{} learns their visual consequences.

These results answer our first question affirmatively: explicitly learning
forward dynamics conditioned on clean actions improves action learning
within the same causal model. Both WAM variants already receive
future-video supervision, so the gains highlight the additional value of
learning how specified controls change the scene. This benefit carries
over to deployment, where the policy generates actions from noise.
The stronger performance across scene perturbations and complex
interactions suggests that clean-action world modeling shapes
representations of robot--scene dynamics that remain useful when
future controls must be inferred, supporting policy generalization
across changing visual configurations.

\FloatBarrier
\subsection{Action Learning for World Modeling}
\label{sec:action-for-world}
We next address our second question: whether action learning improves
action-conditioned world modeling. We compare \model{} with
\wmbaseline{}, which uses the same MoT architecture but learns only
future-video prediction conditioned on clean actions, without an
action-prediction objective. Both receive clean action chunks at inference.
To distinguish joint attention from action modulation, we also design
an IRASim-inspired~\citep{zhu2024irasim} world-model baseline,
\wmadaln{}, that injects action conditions through adaptive layer
normalization (AdaLN). It uses the same Wan2.2-TI2V-5B backbone as our
MoT models. Table~\ref{tab:world-prediction} additionally includes
Wan~\citep{wan2025wan} as a video-prediction reference.

\begin{table}[!htbp]
  \centering
  \captionsetup{skip=4pt}
  \caption{\CaptionLead{World-prediction performance.}
  Arrows indicate the preferred direction. Bold and underline mark the
  best and second-best values within each benchmark, respectively.}
  \label{tab:world-prediction}
  \small
  \setlength{\tabcolsep}{2pt}
  \renewcommand{\arraystretch}{1.02}
  \setlength{\aboverulesep}{1.5pt}
  \setlength{\belowrulesep}{1.5pt}
  \begin{tabular}{@{}ccccc@{\hspace{7pt}}cccc@{\hspace{7pt}}cccc@{}}
    \toprule
    & \multicolumn{4}{c}{\sdtablehead{LIBERO}} & \multicolumn{4}{c}{\sdtablehead{LIBERO-Plus}} & \multicolumn{4}{c}{\sdtablehead{RoboCasa365}} \\
    \cmidrule(lr){2-5}\cmidrule(lr){6-9}\cmidrule(l){10-13}
    \sdtablehead{Metric} & Wan & \wmadaln{} & \wmbaseline{} & \model{} & Wan & \wmadaln{} & \wmbaseline{} & \model{} & Wan & \wmadaln{} & \wmbaseline{} & \model{} \\
    \midrule
    PSNR $\uparrow$ & 20.52 & 20.52 & \underline{22.33} & \textbf{22.95} & 19.08 & 19.58 & \underline{20.54} & \textbf{21.05} & 23.27 & 22.22 & \underline{24.72} & \textbf{25.21} \\
    SSIM $\uparrow$ & 0.8045 & 0.8057 & \underline{0.8325} & \textbf{0.8423} & 0.7575 & 0.7813 & \underline{0.7937} & \textbf{0.8049} & 0.8176 & 0.7873 & \underline{0.8433} & \textbf{0.8515} \\
    LPIPS $\downarrow$ & 0.1817 & 0.1793 & \underline{0.1529} & \textbf{0.1458} & 0.2448 & 0.2185 & \underline{0.2014} & \textbf{0.1935} & 0.1292 & 0.1564 & \underline{0.1087} & \textbf{0.1015} \\
    FID $\downarrow$ & 14.00 & 12.88 & \textbf{11.89} & \underline{12.45} & 24.01 & \textbf{15.91} & 18.06 & \underline{17.01} & \textbf{9.04} & 10.28 & 9.20 & \underline{9.16} \\
    FVD $\downarrow$ & 15.83 & 14.73 & \underline{13.68} & \textbf{13.58} & 49.73 & \textbf{29.53} & 32.48 & \underline{31.66} & 20.23 & \textbf{15.98} & 18.55 & \underline{17.74} \\
    \bottomrule
  \end{tabular}
\end{table}

\paragraph{Action learning strengthens conditional prediction.}
\model{} achieves the best PSNR, SSIM, and LPIPS on all three benchmarks,
with \wmbaseline{} consistently second. Since the two models share the
MoT architecture, this pattern supports the benefit of adding action
learning to world-model training. Predicting actions requires the observed
representation to capture the robot--object configurations relevant to
control. This complementary supervision can strengthen the scene and
forward-dynamics representations used to predict the visual consequences
of specified actions. The improvements across pixel accuracy, structural
similarity, and perceptual correspondence are consistent with this richer
understanding of action-dependent scene changes.

\paragraph{Strong prediction across complementary metrics.}
Compared with \underline{\wmbaseline{} using the same MoT architecture},
\model{} consistently improves paired prediction quality and FVD across
the three benchmarks, as well as FID on LIBERO-Plus and RoboCasa365.
As the architecture-matched comparison, this fact provides a clear evidence
for the benefit of adding action learning.
\model{} ranks first or second on every metric across the three
benchmarks, with the modulation-based \wmadaln{} achieving higher
performance on FID/FVD in some settings---an advantage in matching
feature distributions also observed over the MoT-based \wmbaseline{}.
Paired prediction metrics and distributional metrics characterize
complementary aspects of world modeling: the former measure agreement
with the reference outcome under a given conditioning sequence,
while FID and FVD compare marginal feature distributions without
explicitly evaluating action--outcome correspondence
~\citep{heusel2017fid,unterthiner2018fvd,zhu2024irasim}.
Appendix~\ref{app:qualitative-world-prediction} further evaluates
this correspondence through local geometry and action-dependent
motion in external- and wrist-camera predictions across all
three benchmarks.

These results answer our second question affirmatively: action learning
improves the prediction of specified-action outcomes under identical
clean-action inference conditions. Learning to generate actions gives
the shared model additional supervision for robot--object relationships
relevant to control. This complementary signal can strengthen the scene
representations used to predict the visual consequences of specified
actions. Together with Sec.~\ref{sec:world-for-action},
the results suggest a reciprocal role for the two objectives: forward
prediction learns how controls change a scene, while action generation
emphasizes the scene structure needed to choose them. Both can strengthen
the same representations of interaction under unified causal training.

\subsection{Effect of the Action-Conditioning Mixture}
\label{sec:conditioning-mixture}
The preceding sections examine the two directions of transfer separately.
We now consider how the conditioning mixture in
Sec.~\ref{sec:action-conditioning} affects both tasks. We vary the sampling
probabilities of \emph{all-noisy}, \emph{partial-clean}, and
\emph{all-clean} examples, comparing 60/25/15, 50/35/15, and 30/45/25.
Table~\ref{tab:conditioning-mixture} reports policy success alongside
world-prediction quality; the inference settings remain those in
Sec.~\ref{sec:inference}.

\begin{table}[!htbp]
  \centering
  \begin{minipage}{0.84\textwidth}
  \centering
  \captionsetup{skip=4pt}
  \caption{\CaptionLead{Conditioning mixtures affect both tasks.}
  Columns specify all-noisy / partial-clean / all-clean percentages;
  60/25/15 is the default. SR is success rate (\%). Bold marks the best
  value within each benchmark.}
  \label{tab:conditioning-mixture}
  \small
  \setlength{\tabcolsep}{4pt}
  \renewcommand{\arraystretch}{1.02}
  \setlength{\aboverulesep}{1.5pt}
  \setlength{\belowrulesep}{1.5pt}
  \begin{tabular}{@{}lrrr@{\hspace{12pt}}rrr@{}}
    \toprule
    & \multicolumn{3}{c}{\sdtablehead{LIBERO}} & \multicolumn{3}{c}{\sdtablehead{LIBERO-Plus}} \\
    \cmidrule(lr){2-4}\cmidrule(l){5-7}
    \sdtablehead{Metric} & 60/25/15 & 50/35/15 & 30/45/25 & 60/25/15 & 50/35/15 & 30/45/25 \\
    \midrule
    SR (\%) $\uparrow$ & 94.95 & \textbf{95.95} & 87.80 & \textbf{59.27} & 56.98 & 38.01 \\
    PSNR $\uparrow$ & 22.95 & 22.44 & \textbf{23.12} & \textbf{21.05} & 20.57 & 20.90 \\
    SSIM $\uparrow$ & 0.8423 & 0.8347 & \textbf{0.8443} & \textbf{0.8049} & 0.7943 & 0.7993 \\
    LPIPS $\downarrow$ & 0.1458 & 0.1517 & \textbf{0.1444} & \textbf{0.1935} & 0.2013 & 0.1966 \\
    FID $\downarrow$ & 12.45 & 12.57 & \textbf{12.36} & \textbf{17.01} & 17.32 & 17.18 \\
    FVD $\downarrow$ & \textbf{13.58} & 14.64 & 13.97 & 31.66 & 33.18 & \textbf{31.56} \\
    \bottomrule
  \end{tabular}
  \end{minipage}
\end{table}

Policy performance and world-prediction quality remain broadly stable
between 60/25/15 and 50/35/15. The modest shift toward partial-clean
conditioning brings a small in-distribution policy gain, while the
default remains stronger on LIBERO-Plus and paired prediction metrics.
Joint visual and action learning is therefore reasonably stable to
moderate changes in the mixture.

A larger shift toward clean conditioning leaves world-prediction metrics
relatively stable but weakens policy performance, particularly under
distribution shift. This pattern suggests that policy generalization
benefits from preserving sufficient all-noisy training, which matches
deployment where actions are generated from noise. Clean-action
conditioning supplies forward-dynamics supervision, but increasing
its share favors reconstruction in familiar scenes without improving
prediction on LIBERO-Plus. The two learning signals therefore need to
be balanced: more clean conditioning alone does not ensure better
generalization. Among the tested mixtures, 60/25/15 provides the best
balance of policy generalization and paired world-prediction quality.

\subsection{Endless Dream: Closing the Loop in Imagination}
\label{sec:endless-dream}
Beyond the improvements above, the two inference modes in
Sec.~\ref{sec:inference} can compose into an \emph{endless dream}: \model{}
alternates action generation and action-conditioned world prediction,
using each imagined outcome as the context for its next decision.
Let $\hat{\mathbf{o}}_0$ be the initial multi-view observation and
$\mathcal{C}_j=(E(\hat{\mathbf{o}}_j),\ell)$ the context at round $j$,
where $E$ and $D$ denote the video VAE encoder and decoder. The loop is
\begin{equation}
  \begin{aligned}
  \hat{\mathbf{a}}_j &\sim p_\theta(\mathbf{a}\mid\mathcal{C}_j),
  &\hat{\mathbf{z}}^{\mathrm{fut}}_j
    &\sim p_\theta(\mathbf{z}^{\mathrm{fut}}
      \mid\mathcal{C}_j,\hat{\mathbf{a}}_j),\\
  \hat{\mathbf{o}}_{j+1}
    &= \operatorname{last}\!\left(D(\hat{\mathbf{z}}^{\mathrm{fut}}_j)\right).
  \end{aligned}
  \label{eq:endless-dream}
\end{equation}
Both stages use the same model: the action chunk is first denoised from
noise, then held fixed as clean conditioning for video generation. The
final predicted frame from each camera is fed back into the next round,
with the instruction unchanged and no further environment observations.

On LIBERO tasks, we run this loop for 16 rounds without additional
training. Each round generates 32 actions and eight future frames,
yielding 512 actions over a modeled horizon of 25.6 seconds. The resulting
sequences show task-directed progression through drawer opening, pushing,
and single- and multi-object transport
(Appendix~\ref{app:endless-dream}). This composability connects the two
learned capabilities beyond separate evaluation: action generation selects
controls from imagined states, while forward prediction supplies the
states on which subsequent decisions depend.

\begin{samepage}
\section{Conclusion}
\label{sec:conclusion}
We studied whether world and action modeling can benefit each other under
unified causal training. \model{} combines clean-action-conditioned forward
prediction and action generation in a causal MoT model by mixing all-clean,
partial-clean, and all-noisy examples. Controlled simulation comparisons
provide positive answers to both questions: explicit forward-dynamics
learning improves policy performance and generalization, while action
learning improves action-conditioned world prediction within the same
architecture. Real-world manipulation further validates the policy
benefit. These results suggest that the two objectives supervise
complementary aspects of interaction: the visual consequences of controls
and the scene structure needed to select them. The mixture study shows
that this benefit tolerates moderate rebalancing, while larger shifts
toward clean conditioning mainly weaken policy generalization.
\model{} thus supports learning world prediction and control through
mutually beneficial representations under temporal causality.
These findings motivate studying how reciprocal learning scales with more
diverse data and longer interaction horizons. A further direction is to
use the learned dynamics for planning and policy improvement, extending
the interaction between prediction and control beyond joint training.

\end{samepage}

\clearpage
\bibliographystyle{unsrtnat}
\bibliography{references}

\clearpage
\appendix
\section{Training and Evaluation Protocols}
\label{app:evaluation-protocol}

\subsection{Training and Deployment}
\label{app:training-deployment}

\paragraph{Simulation training.}
For LIBERO and LIBERO-Plus, we adopt the LIBERO training recipe of
Fast-WAM~\citep{yuan2026fastwam}. We train jointly on demonstrations
from the Spatial, Object, Goal, and Long suites for 20 epochs with a
global batch size of 128. The resulting models are evaluated on
LIBERO-Plus directly. For RoboCasa365, we train
on its demonstration dataset for 90k optimizer steps with a global
batch size of 1024.

\paragraph{Real-world training.}
We collect 50 demonstrations for each of the three real-world tasks:
block stacking, block insertion, and breakfast arrangement, yielding
150 episodes in total. Models are trained on these demonstrations for
20 epochs with a global batch size of 96.

\paragraph{Internal comparisons.}
All internal WAM and WM ablations follow the same dataset-specific
training protocol, including training data, global batch size, and
training budget. Their action conditioning and learning objectives
differ as described in Sec.~\ref{sec:experimental-setup}.
Unless otherwise stated, \model{} samples all-noisy, partial-clean,
and all-clean action conditioning with probabilities $60\%$, $25\%$,
and $15\%$, respectively.

\paragraph{Deployment.}
Following Sec.~\ref{sec:inference}, policies generate action chunks
from Gaussian noise using the current visual observations and language
instruction. Observed video representations are reused during action
denoising, without generating future video. For world prediction, the
specified action chunk remains clean and fixed while the model denoises
future video latents.

\subsection{Evaluation Protocols}

\paragraph{Policy evaluation.}
We measure closed-loop task success using each benchmark's success
criterion. LIBERO results cover the Spatial, Object, Goal, and Long
suites and their average; LIBERO-Plus results cover the seven perturbation
categories and the benchmark aggregate. RoboCasa365 results are reported
for atomic-seen, composite-seen, and composite-unseen tasks. Each
real-world task is evaluated over 15 trials per model, with success rates
reported as percentages. Inference follows Sec.~\ref{sec:inference}.

\paragraph{Video evaluation.}
LIBERO, LIBERO-Plus, and RoboCasa365 are designed for policy evaluation
and do not provide dedicated validation sets for action-conditioned video
prediction. We therefore construct evaluation video sets from expert
rollouts. For LIBERO and LIBERO-Plus, we collect 100 complete episodes in
total using $\pi_{0.5}$~\citep{intelligence2025pi05}, uniformly sampling
across LIBERO settings and LIBERO-Plus perturbation categories. For
RoboCasa365, we collect 100 complete episodes with
Xiaomi-Robotics-1~\citep{xiaomi2026robotics1}, uniformly sampling its
50 evaluation target tasks. All three benchmarks are evaluated with multi-view future predictions,
including external-camera and wrist-camera observations. Prediction
chunks span 32 action steps with
a frame-sampling stride of 4 for LIBERO and LIBERO-Plus, and 16 action
steps with a stride of 2 for RoboCasa365. Each chunk thus evaluates eight
future frames, excluding the conditioning frames.
PSNR, SSIM, and LPIPS are computed from losslessly stored RGB frames
scaled to $[0,1]$ using PIQ~0.8~\citep{kastryulin2022piq}. For FID and FVD,
both predictions and references are encoded with H.264 at constant
quantization parameter 20 in YUV420p format and decoded before feature
extraction.

\paragraph{PSNR.}
Peak signal-to-noise ratio measures pixel reconstruction accuracy as
$-10\log_{10}(\mathrm{MSE}+10^{-8})$ in decibels, where MSE is averaged
over RGB channels and pixels. Higher is better.

\paragraph{SSIM.}
Structural similarity~\citep{wang2004ssim} compares local luminance,
contrast, and structure. We use PIQ's single-scale implementation with
an $11\times11$ Gaussian window, standard deviation $1.5$, unit data
range, and $K_1=0.01$, $K_2=0.03$. Higher is better.

\paragraph{LPIPS.}
Learned perceptual image patch similarity~\citep{zhang2018lpips} measures
perceptual distance using PIQ's ImageNet-pretrained VGG16 backbone,
learned calibration weights, and default input normalization.
Lower is better.

\paragraph{FID.}
Fr\'{e}chet Inception Distance~\citep{heusel2017fid} compares generated
and reference image-feature distributions using 2,048-dimensional
average-pooling features from PIQ's TensorFlow-compatible Inception-v3.
Frames are bilinearly resized to $299\times299$ and normalized to
$[-1,1]$. Lower is better.

\paragraph{FVD.}
Our Fr\'{e}chet Video Distance~\citep{unterthiner2018fvd} implementation
uses 400-dimensional output logits from PyTorchVideo's I3D-R50 pretrained
on Kinetics-400. Each clip contains eight future frames in temporal order,
bilinearly resized to $224\times224$ and normalized from $[0,1]$ with
channel-wise mean $0.45$ and standard deviation $0.225$. Lower is better.

PSNR, SSIM, and LPIPS are averaged over all evaluated future frames.
FID and FVD use feature means and covariances pooled over each benchmark
to compute the Fr\'{e}chet distance, rather than averaging per-batch scores.

\clearpage
\section{Qualitative Analysis of Action-Conditioned World Prediction}
\label{app:qualitative-world-prediction}
We inspect selected, matched prediction windows from the evaluated model
variants on LIBERO, LIBERO-Plus, and RoboCasa365. Each comparison uses the same observed
context, instruction, clean action sequence, and camera. The figures show
four future frames and a shared-coordinate detail at the final time step.

\begin{figure}[!htbp]
  \centering
  \includegraphics[width=\textwidth]{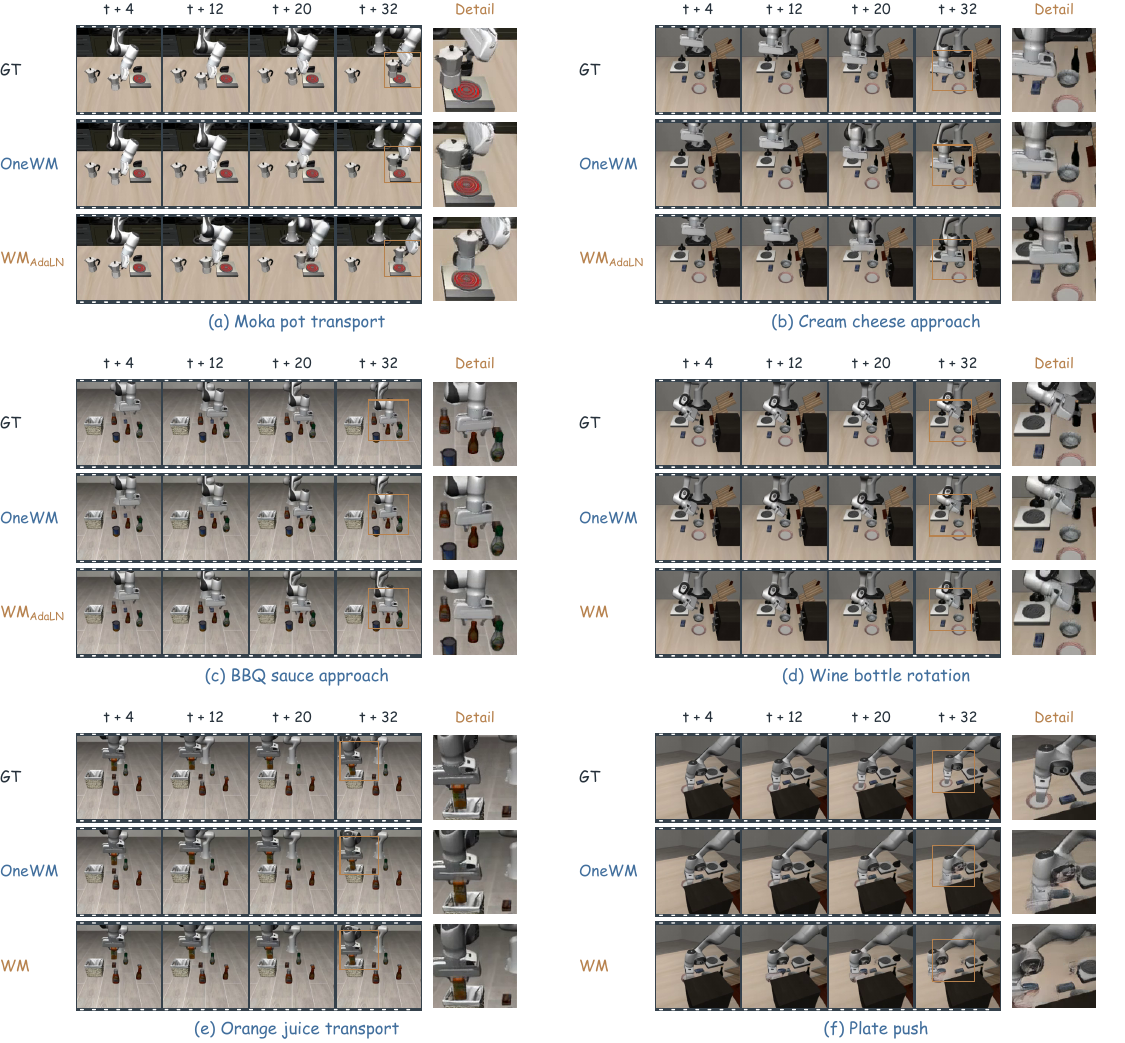}
  \caption{\CaptionLead{Main-camera world predictions.}
  Six cases from LIBERO and LIBERO-Plus compare GT, \model{}, and the
  indicated world-model baseline. Orange boxes mark the regions enlarged
  at $t+32$ in each panel's final column.}
  \label{fig:qualitative-main}
\end{figure}

\paragraph{Main-camera predictions.}
Figure~\ref{fig:qualitative-main} highlights spatial relationships at the
scene level. Compared with \wmadaln{}, \model{} more closely follows
the held moka pot and gripper pose in (a), the gripper's position relative
to the cream cheese box in (b), and the gripper height near the sauce
bottle in (c). Comparisons with \wmbaseline{} show closer correspondence
in the bottle rotation (d), carton--basket alignment (e), and gripper--plate
relationship (f). These examples illustrate the value of action learning
for representing how robot motion changes the scene configuration.

\begin{figure}[!htbp]
  \centering
  \includegraphics[width=\textwidth]{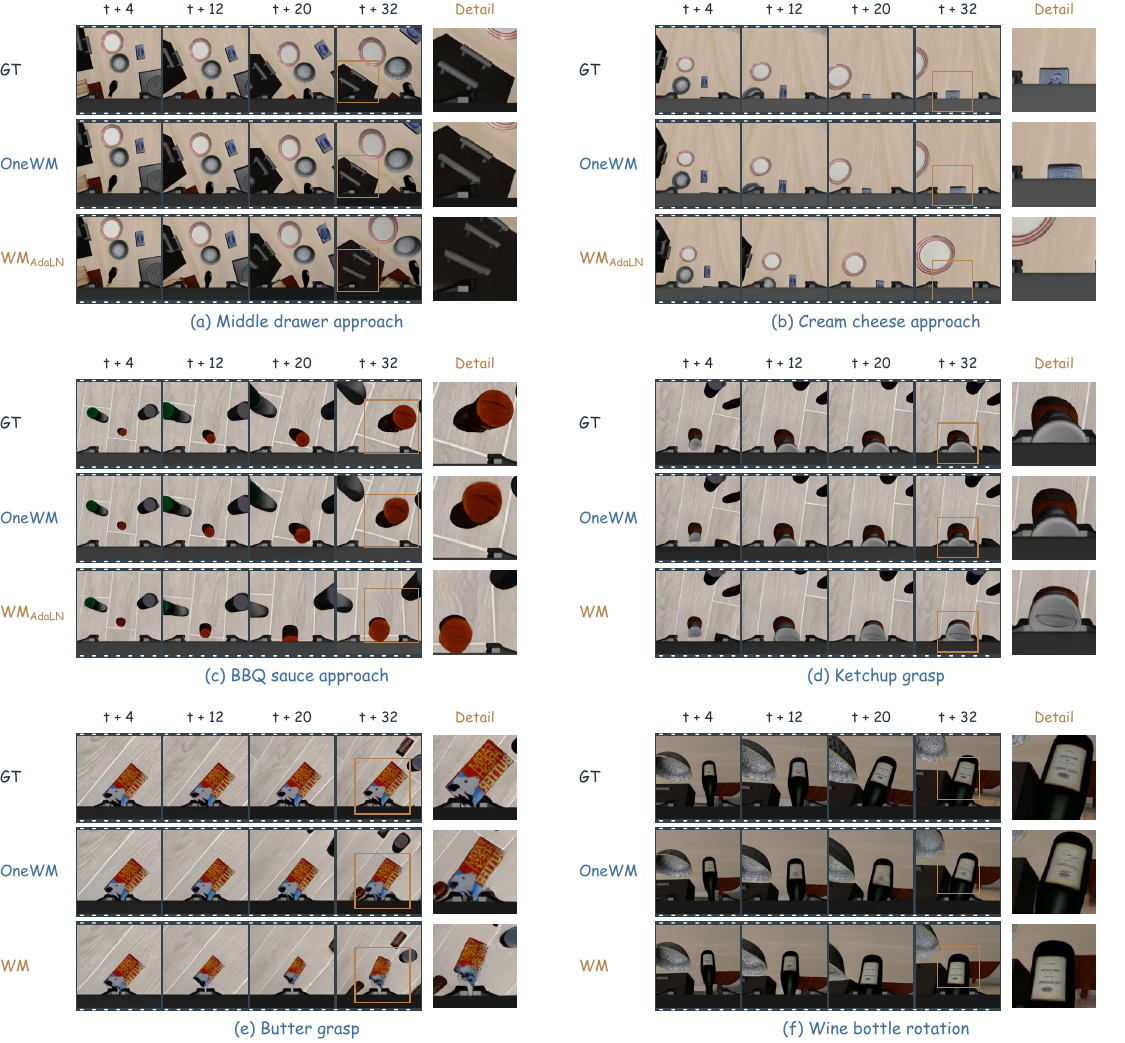}
  \caption{\CaptionLead{Wrist-camera world predictions.}
  Six matched comparisons emphasize object position, shape, and pose near
  the gripper. Each final column enlarges the same marked region at
  $t+32$ for all three rows.}
  \label{fig:qualitative-wrist}
\end{figure}

\clearpage
\paragraph{Wrist-camera predictions.}
Figure~\ref{fig:qualitative-wrist} reveals local geometry during interaction.
Against \wmadaln{}, \model{} better preserves the drawer's image position
in (a), the cream cheese box near the gripper in (b), and the sauce cap's
position and orientation in (c). Against \wmbaseline{}, it better retains
the visible bottle-neck geometry in (d), the butter package's extent in
(e), and the bottle inclination in (f). The cream cheese, sauce, and wine
cases also appear in the main view, allowing the same predictions to be
inspected from complementary cameras.

\paragraph{Conditional fidelity and distributional quality.}
These examples help interpret the different metric rankings in
Table~\ref{tab:world-prediction}. PSNR, SSIM, and LPIPS compare each
prediction with its corresponding reference, whereas FID and FVD compare
pooled feature statistics~\citep{heusel2017fid,unterthiner2018fvd}.
Reassigning generated clips to different action sequences would leave
these pooled statistics unchanged. A lower distributional distance
therefore need not imply more accurate action-conditioned geometry.

\begin{figure}[!htbp]
\centering
\includegraphics[width=0.98\textwidth]{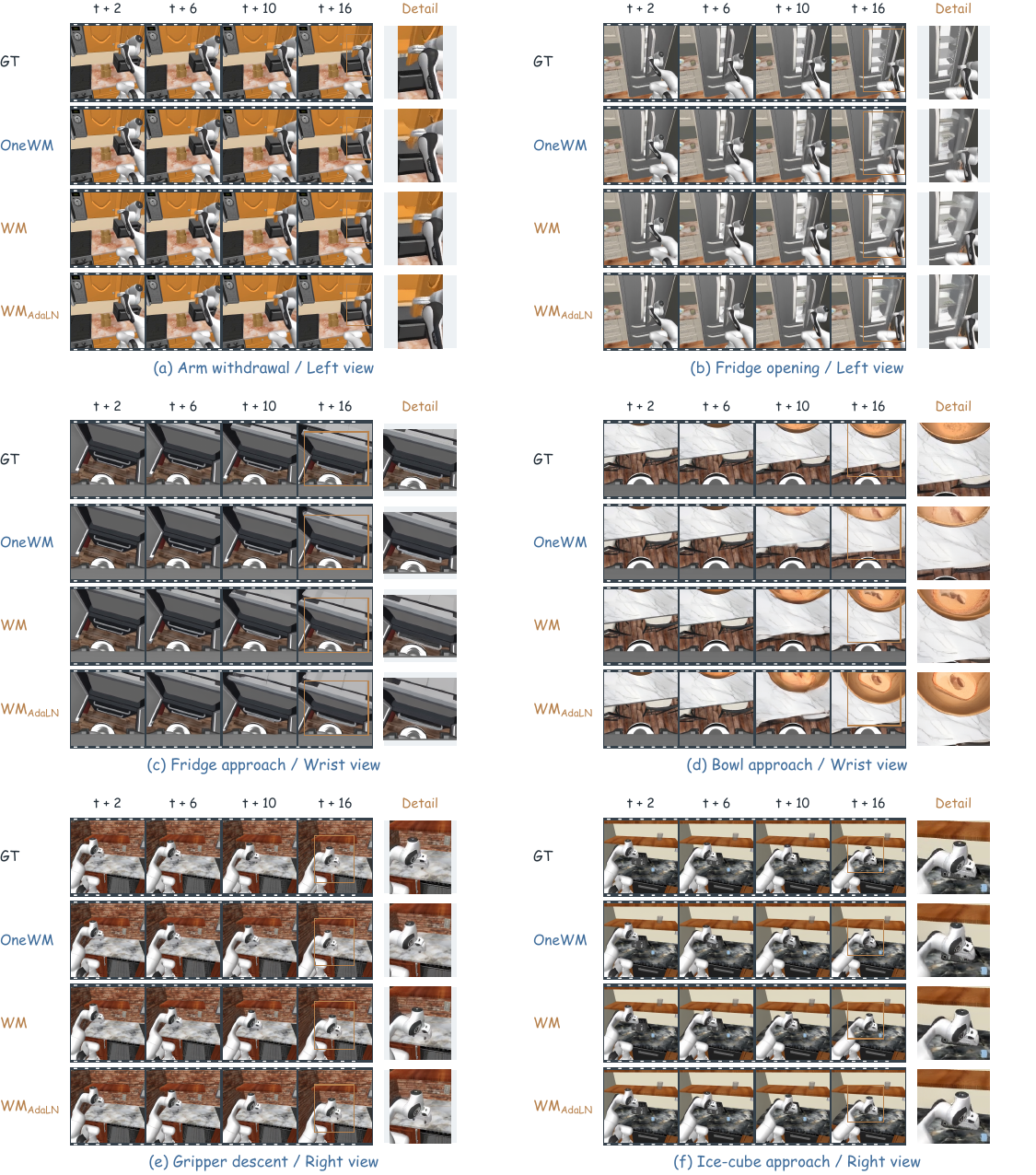}
\caption{\textbf{RoboCasa365 world predictions.} Six matched cases
compare GT, \model{}, \wmbaseline{}, and \wmadaln{} from left (a,b),
wrist (c,d), and right (e,f) cameras. The final column enlarges the
same marked region at $t+16$ for all four rows.}
\label{fig:qualitative-robocasa}
\end{figure}

\clearpage
\paragraph{RoboCasa365 predictions.}
Figure~\ref{fig:qualitative-robocasa} extends the comparisons to diverse
scenes across three camera views. In the left views (a,b), \model{}
more closely tracks forearm withdrawal and gripper--door motion.
In the wrist views (c,d), the baselines overpredict viewpoint change,
shifting the fridge geometry and bringing the bowl into view too early.
The right views (e,f) show closer correspondence in gripper descent
and orientation near the counter and ice cube. These differences
concern motion relative to scene structure, rather than appearance
alone, illustrating the value of action learning for conditional
spatial prediction.

\clearpage
\section{Endless Dream on LIBERO}
\label{app:endless-dream}
\begin{figure}[!htbp]
  \centering
  \includegraphics[width=0.98\textwidth]{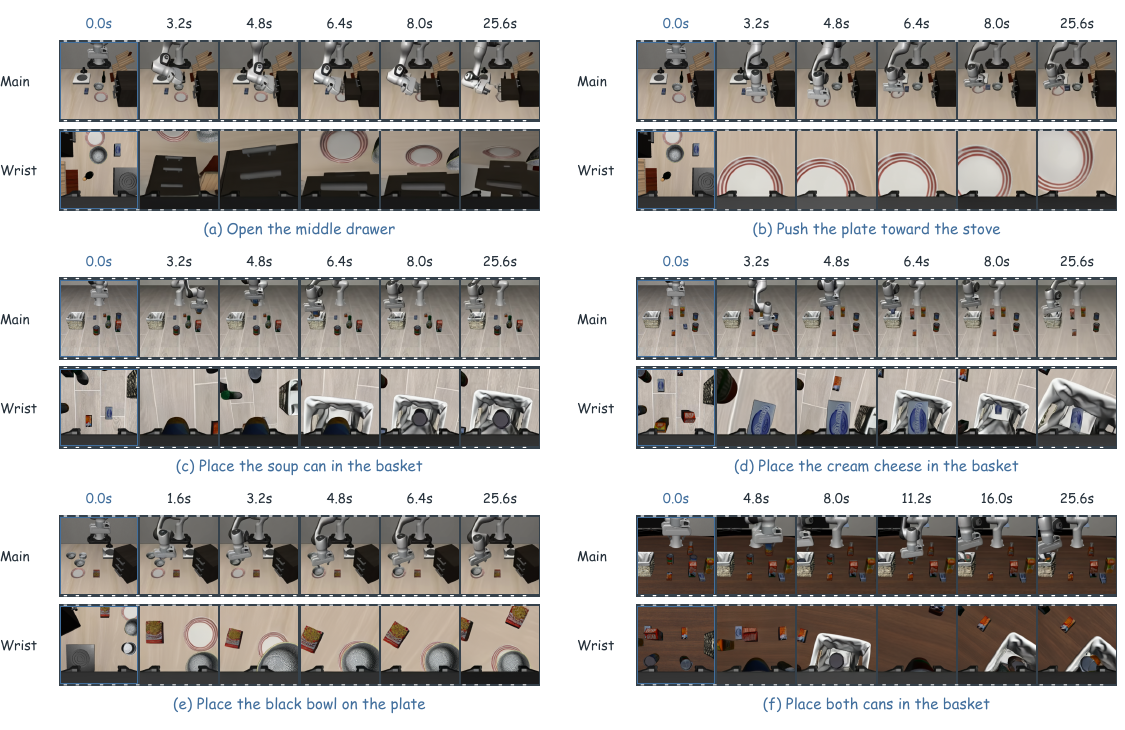}
  \caption{\CaptionLead{Endless dream on LIBERO.}
  Six rollouts generated by alternating action prediction and
  action-conditioned world prediction. Each panel shows main- and
  wrist-camera views; blue outlines mark the initial observations,
  and every later frame is generated. Columns indicate modeled time,
  with the final column showing the end of round 16.}
  \label{fig:endless-dream}
\end{figure}

We instantiate the loop in Eq.~\ref{eq:endless-dream} with the jointly
trained LIBERO model and the default conditioning mixture. Each rollout
starts from one main-camera image, one wrist-camera image, and a task
instruction. Both inference stages use 10 denoising steps with a fixed
seed of 42. The generated action chunk is passed unchanged to world
prediction; its final multi-view prediction becomes the next visual
context. Only the initial observation is obtained from the environment.
All rollouts use a fixed budget of 16 rounds, with 32 actions and
eight predicted frames per round. Figure~\ref{fig:endless-dream} shows
intermediate states and the final state of every rollout.

\end{document}